\documentclass[letterpaper, 10pt, conference]{ieeeconf}
\IEEEoverridecommandlockouts

\usepackage{cite}
\usepackage{amsmath, amssymb, amsfonts, mathtools}
\usepackage{algorithmic}
\usepackage[short]{optidef}
\usepackage{graphicx}
\usepackage{textcomp}
\usepackage{xcolor}
\usepackage{booktabs}
\usepackage{multirow}
\usepackage{tikz}

\newcommand\copyrighttext{
	\footnotesize \textcopyright 2023 IEEE. Personal use of this material is permitted. Permission from IEEE must be obtained for all other uses, in any current or future media, including reprinting/republishing this material for advertising or promotional purposes, creating new collective works, for resale or redistribution to servers or lists, or reuse of any copyrighted component of this work in other works.}
\newcommand\copyrightnotice{
	\begin{tikzpicture}[remember picture,overlay]
		\node[anchor=south,yshift=5pt] at (current page.south) {\fbox{\parbox{\dimexpr\textwidth-\fboxsep-\fboxrule\relax}{\copyrighttext}}};
	\end{tikzpicture}
}

\DeclareMathSymbol{\shortminus}{\mathbin}{AMSa}{"39}

\newtheorem{assumption}{Assumption}

\newtheorem{lemma}{Lemma}
\newcommand{\ra}[1]{\renewcommand{\arraystretch}{#1}}
\newcommand{\E}{\mathbb{E}}
\newcommand{\R}{\mathbb{R}}

\renewcommand{\arraystretch}{1.2}
\def\BibTeX{{\rm B\kern-.05em{\sc i\kern-.025em b}\kern-.08em
T\kern-.1667em\lower.7ex\hbox{E}\kern-.125emX}}

\begin{document}

\title{\bf Imitation Learning from Nonlinear MPC via the Exact Q-Loss and its Gauss-Newton Approximation}
\author{{Andrea Ghezzi$^{\star, 1}$, Jasper Hoffman$^{\star, 2}$, Jonathan Frey$^{1,3}$, Joschka Boedecker$^2$,  Moritz Diehl$^{1,3}$}
\thanks{$^\star$ These authors contributed equally.}
\thanks{$^1$ Department of Microsystems Engineering (IMTEK), University of Freiburg, 79110 Freiburg, Germany}
\thanks{$^2$ Department of Computer Science, University of Freiburg}
\thanks{$^3$ Department of Mathematics, University of Freiburg, 79104 Freiburg, Germany}
\thanks{Corresponding authors: \{andrea.ghezzi@imtek.uni-freiburg.de, hoffmaja@informatik.uni-freiburg.de\}}
\thanks{This research was supported by DFG via Research Unit FOR 2401 and project 424107692 and by the EU via ELO-X 953348.}}
\maketitle

\copyrightnotice

\begin{abstract}

    This work presents a novel loss function for learning nonlinear Model Predictive Control policies via Imitation Learning.
    Standard approaches to Imitation Learning neglect information about the expert and generally adopt a loss function based on the distance between expert and learned controls.
    In this work, we present a loss based on the Q-function directly embedding the performance objectives and constraint satisfaction of the associated Optimal Control Problem (OCP).
    However, training a Neural Network with the Q-loss requires solving the associated OCP for each new sample.
    To alleviate the computational burden, we derive a second Q-loss based on the Gauss-Newton approximation of the OCP resulting in a faster training time.
    We validate our losses against Behavioral Cloning, the standard approach to Imitation Learning, on the control of a nonlinear system with constraints.
    The final results show that the Q-function-based losses significantly reduce the amount of constraint violations while achieving comparable or better closed-loop costs.

\end{abstract}

\section{Introduction}
Model Predictive Control (MPC) is an optimization-based approach for controlling dynamical systems \cite{Rawlings2017}.
It is used in many different applications due to the versatility, handling of constraints and stability guarantees.
For each new state, MPC computes a control by solving an optimal control problem (OCP), that contains the performance objectives, system dynamics and system properties.
Yet, the computational complexity of real time optimization can be a major limitation of the applicability of MPC, when a certain control frequency is necessary or one uses embedded systems with constrained computationally resources.

One approach to tackle this problem is \textit{explicit} MPC.
Instead of having an \textit{implicit} control policy that is derived from online optimization, \textit{explicit} MPC computes the control policy beforehand in an offline setting which then can be simply evaluated online.
For linear MPC it is possible to represent the state feedback policy by a piece-wise affine function and store the corresponding gain as a look-up table.
However, for nonlinear MPC (NMPC) is not possible to exactly represent the policy and the available approaches suffer from approximation errors.
Therefore, the common choice is to rely on the implicit online computation of the control policy.

A promising approach related to \textit{explicit} MPC is Imitation Learning (IL).
Here, one tries to imitate the behavior of the MPC policy with a parameterized policy like a Neural Network.
The main issue of this approach is that we introduce approximation errors while imitating.
On the other hand, IL can be used, even in the case of NMPC, to drastically reduce online computational costs.
Previous works mainly used methods like Behavioral Cloning (BC) to imitate the MPC policy.
However, the loss function used in BC is just a surrogate loss that minimizes the difference between the MPC policy and the learned policy \cite{Osa2018}.
With such a surrogate loss, we lose all information why the MPC took a control in the first place.
Thus, during training, the learned policy gets no feedback in terms of constraint satisfaction and performance objectives.

In this work, instead of falling back to a surrogate loss like in BC, we introduce a new loss formulation that exposes to the policy the inner objective of the MPC during training.
For this, we introduce a Q-function which corresponds to the cost at the optimal solution of the associated OCP for a given initial state and fixed initial control.

Specifically, our contributions are twofold:
\begin{enumerate}
    \item we propose a loss for IL based on the Q-function of the given OCP such that the loss directly embeds the characteristics of the OCP;
    \item we introduce a quadratic programming approximation of the Q-function loss to reduce the computational burden and make it more suitable for training Neural Networks.
\end{enumerate}
We compare both proposed loss functions against Behavioral Cloning (BC) on the stabilization of a nonlinear cart-pole system.
The policies learned with the Q-function losses achieve a lower cost and significantly less constraint violations compared to the BC policy.
This is promising since we expect that the potential performance difference between the proposed losses and the standard one will be even more evident on more complex examples.

\paragraph*{Related work}
For \textit{explicit} linear MPC, it is possible to have an exact representation of the policy via piece-wise affine functions \cite{Bemporad1999, Bemporad2002b}.
Via approximation of such policies, \textit{explicit} MPC has been extended to the nonlinear case \cite{Johansen2004}.
The use of Neural Networks to represent optimal control policies has been investigated in \cite{Parisini1995} and applied to chemical process control in \cite{Aakesson2006}.
Thanks to the success of deep learning, the use of Neural Network for \textit{explicit} MPC has grown, with recent applications in power electronics, building control and robot manipulators \cite{Lucia2020, Drgovna2021, Nubert2020}.

Many works have started to investigate properties and guarantees for these learned controllers, in \cite{Chen2018b} an approximate linear MPC is trained via Reinforcement Learning with guaranteed feasible controls obtained by projection on the control invariant set of the system.
For NMPC, in \cite{Hertneck2018} a statistical guarantee on stability and constraint satisfaction is derived via a condition on the approximation error of the learned MPC.
In \cite{Cosner2022}, control barrier functions are introduced as a way to transfer safety from the expert to the learned controller and a formal guarantee of input-to-state stability is provided.
In \cite{Mordatch2014} a trajectory is not only optimized for costs but also whether the trajectories can be recreated by the learned policy.
However, the approaches mentioned above are based on Behavior Cloning, thus they minimize a surrogate loss function.

A very related line of work, looking at IL learning from MPC, can be found in \cite{Carius2020} and \cite{Reske2021}, where IL is reformulated by minimizing the control Hamiltonian of a continuous-time OCP formulation.
One major difference to our approach, apart from that we look at discrete-time OCPs, is that with the Hamiltonian, one does not resolve the OCP for each new control, but only use the gradient of the cost-to-go function with respect to the state.
In order to incorporate information about inequality constraints into this gradient, they introduce log-barrier functions to the OCP formulation.

\subsection{Notation}
Given $a \in \mathbb{R}^{n_a}$ and $b \in \mathbb{R}^{n_b}$, we denote the vector $c = [a^\top \; b^\top]^\top$ by $c = (a, b)$.
Given $a \in \mathbb{R}^{n_a}$, we denote $\Vert a \Vert_W^2 \coloneqq a^\top W a$, for every $W \in \mathbb{R}^{n_a \times n_a}$ a positive definite matrix.
With $\mathcal{U}(a, b)$, we denote the uniform distribution with boundaries $a, b$ respectively.

\section{Background}
In this section, we provide the necessary background of Optimal Control and Imitation Learning.

\subsection{Optimal Control}
In this work, we want to approximate NMPC policies, which are defined by the repetitive solution of an OCP.
Specifically throughout this paper, we regard the following generic discrete-time OCP
\begin{mini!}[2]
    {\substack{x_0, u_0, s_0, \dots, \\ u_{N-1}, x_N, s_N}}{\sum_{k=0}^{N-1}\tilde{L}(x_k, u_k, s_k) + \tilde{E}(x_N, s_N)}{\label{ocp: discrete-time OCP}}{\label{cf: discrete-time OCP}}
    \addConstraint{x_0}{= \bar{x}_0\label{cns: initial state constraint}}
    \addConstraint{x_{k+1}}{= f(x_k, u_k),}{\, k=0,\dots, N-1  \label{cns: system dynamic}}
    \addConstraint{h(x_k, u_k)}{\leq s_k,}{\, k=0,\dots, N-1 \label{cns: h<=0}}
    \addConstraint{r(x_N)}{\leq s_N \label{cns: r_N<=0}}
    \addConstraint{s_k}{\geq 0,}{\, k=0,\dots, N,}
\end{mini!}
with $N$ shooting intervals.
Here, $x_k\in\mathbb{R}^{n_x}$ and $u_k\in\mathbb{R}^{n_u}$ represent the state and the control trajectories which follow the possibly nonlinear system dynamics $f$ in \eqref{cns: system dynamic}.
In order to guarantee feasibility, we introduce slack variables $s_k \in \R^{n_{s,k}}$ and we penalize their use in the cost function.
Thus, the stage cost is defined as $\tilde{L}(x_k, u_k, s_k) \coloneqq L(x_k, u_k) + z^\top s_k + \Vert s_k \Vert_{Z}^2$ and terminal cost as $\tilde{E}(x_N, s_N) \coloneqq E(x_N) +  z_e^\top s_N + \Vert s_N \Vert_{Z_e}^2$.
Inequality \eqref{cns: h<=0} enforces path constraints and \eqref{cns: r_N<=0} encodes a condition on the system terminal state.
We assume the functions $L, E, f, h, r$ to be twice continuously differentiable in their respective variables.
Note, that for some values of the positive slack penalties $z, Z$ an exact penalization of the constraints can be achieved~\cite{Byrd2008}.
In contexts where constraint satisfaction is critical we can tune the weights of the slacks to favor feasibility over optimality.
In an NMPC scheme, the OCP \eqref{ocp: discrete-time OCP} is solved in every control step for a new initial state $\bar{x}_0$ and the optimal control $u_0^\star(\bar{x}_0)$ is applied to the system.

Note that for notational convenience we derive every further OCP formulation by omitting the slack variables.

\subsection{Imitation Learning}
We are interested in using Imitation Learning~(IL) to imitate the control law derived by solving a discrete-time OCP as described in \eqref{ocp: discrete-time OCP}.
We can define the expert policy, that we want to imitate, by the first optimal control $\pi^\star(x) \coloneqq u_0^\star(\bar{x}_0)$ of the solution for a given $\bar{x}_0$.
We aim to approximate $\pi^\star$ as well as possible by a parameterized policy $\pi (\cdot ; w): \R^{n_x} \to R^{n_u}$. 
A parameterized policy could be for example a Neural Network where the parameter $w \in \mathbb{R}^{n_w}$ are the weights of the Neural Network.

In the following, we will give a short introduction to the IL framework.
The IL objective can be defined as
\begin{align}
    \mathcal{L}(w) \coloneqq \E_{x \sim \mathcal{D}} \left[ \ell\left(x, \pi ( \cdot\; ; w)\right)\right],
\end{align}
where $\mathcal{D}$ is a given state distribution over $\R^{n_x}$ and $\ell$ the point wise loss function of the policy $\pi (\cdot\;; w)$ for a given state $x$ \cite{Osa2018}.
The final goal of IL is then to find the optimal combination of parameters $w^\star$ that minimizes the expected loss $\mathcal{L}(w)$:
\begin{argmini}[3]
    {w}{\mathcal{L}(w).}{\label{op: training with L2 loss}}{w^\star = }
\end{argmini}

In its most general form, IL assumes no prior knowledge about the internal objective of the expert policy.
For example the expert could be a human.
Thus, methods like Behavior Cloning, replace the internal objective by using a surrogate loss function $\ell$ that measures the behavioral difference between the policy $\pi$ and the expert policy $\pi^\star$.
Popular choices are the quadratic loss function $\ell_2$ defined as
\begin{align}
    \ell_2(x, \pi) \coloneqq \left( \pi(x) - \pi^\star(x) \right)^2,
\end{align}
the Huber loss \cite{Huber1964}, the $\ell_1$ loss or the cross-entropy loss in the case of stochastic policies.
In this paper, we will use the quadratic loss function $\ell_2$ for comparison, which results in the following expected quadratic loss:
\begin{align}
    \mathcal{L}^2(w) \coloneqq \E_{x \sim \mathcal{D}} \left[ \ell_2\left(x, \pi ( \cdot\; ; w)\right)\right].
\end{align}

\section{Q-loss for Imitation Learning}
In the following, we will see that when the expert policy is the solution of an OCP, we do not necessarily need a surrogate loss.

\subsection{The exact Q-loss}
In fact, we can directly define a loss based on the internal objective of the expert, which we call the exact Q-loss.
This proposed loss function directly embeds the information contained in the OCP such as its cost and constraints.
The main idea is to reuse the OCP formulation~\eqref{ocp: discrete-time OCP} and fix the first control $u_0$ of the OCP by the value returned from the policy, $\bar{u}_0 = \pi(x; w)$.
By solving the resulting OCP, we can assign a cost to the policy value $\pi(x; w)$.

Specifically, we reformulate the OCP~\eqref{ocp: discrete-time OCP} to expose exclusively the first control.
Given $\bar{x}_0$ we define the exact Q-loss by the following ``Q-function OCP''
\begin{mini!}[3]
    {\substack{x_0, u_0, \dots, \\ u_{N-1}, x_N}}{\sum_{k=0}^{N-1} L(x_k, u_k) + E(x_N)}{\label{ocp: q-function ocp}}{Q(\bar{x}_0, \bar{u}_0) \coloneqq}
    \addConstraint{x_0 - \bar{x}_0}{= 0\label{cns: q-ocp x_0=x_0}}
    \addConstraint{u_0 - \bar{u}_0}{= 0 \label{cns: q-ocp u_0=u_0}}
    \addConstraint{x_{k+1} - f(x_k, u_k)}{= 0\label{cns: q-ocp: dynamics constraint}}
    \addConstraint{h(x_k, u_k)}{\leq 0,\,k=1,\dots,N-1}{}
    \addConstraint{r(x_N)}{\leq 0.}
\end{mini!}
We remind that for notational convenience we omit the slack variables in the OCP formulation.
With the exact Q-loss, the imitation learning objective becomes
\begin{align}\label{eq: exact q-loss}
    \mathcal{L}^\mathrm{Q}(w) \coloneqq \E_{x \sim \mathcal{D}} \left[ Q(x, \pi(x; w)) \right].
\end{align}
The name ``Q-loss'' is motivated from the related concept of Q-functions in reinforcement learning.
The gradient of $\mathcal{L}^\mathrm{Q}(w)$ is defined as
\begin{align}\label{eq: gradient L_Q}
    \nabla_w \mathcal{L}^\mathrm{Q}(w) = \E_{x \sim \mathcal{D}}\! \left[ \nabla_w \pi(x; w) \left.\nabla_u Q(x, u) \right|_{u = \pi(x; w)} \right]\!.
\end{align}

\begin{lemma}\label{lemma: lagrangian multiplier}
The gradient of the Q-loss is given by the Langrangian multiplier $\bar{\lambda}_u$ corresponding to the constraint \eqref{cns: q-ocp u_0=u_0} for the optimal solution
\begin{align}
    \bar{\lambda}_u &= \left.\nabla_u Q(x, u) \right|_{u = \pi(x; w)}.
\end{align}
\end{lemma}
\textit{Proof:}
This can be derived by looking at the cost-to-go from dynamic programming and the first order necessary condition of optimality,~\cite[§8.8.3]{Rawlings2017}, \cite[§3.3.3]{Bertsekas2012}. $\square$ \smallskip

\begin{lemma}
    If $\pi^\star(\bar{x}_0)$ is a unique minimizer of \eqref{ocp: discrete-time OCP} then $Q(\bar{x}_0, \bar{u}_0) > Q(\bar{x}_0, \pi^\star(\bar{x}_0))$ for any $\bar{u}_0 \neq \pi^\star(\bar{x}_0)$.
    Thus, the exact Q-loss penalizes any deviation of $\bar{u}_0$ from $\pi^\star(\bar{x}_0)$.
\end{lemma}

When looking at \eqref{eq: exact q-loss}, an interesting connection to actor-critic algorithms \cite{Barto1983} emerges, where the actor is the policy~$\pi$ which is criticized by a parameterized value function $Q$, called the critic.
In this framework, the Q-function derived from the OCP can be seen as the critic.
Following this perspective, the gradient of \eqref{eq: gradient L_Q} is directly related to deterministic policy gradients \cite{Silver2014}.

\subsection{Discussion of the exact Q-loss}\label{subsec: Shortcomings of the exact Q-loss}
The exact Q-loss is computationally more complex than a standard loss.
To get a better understanding of the computational costs, one can compare the gradients of behavior cloning and the exact Q-loss.
For the behavior cloning loss $\mathcal{L}^2$, the gradient is given by
\begin{align}\label{eq: gradient L2 loss}
	\nabla_w \mathcal{L}^2 (w) = \E_{x \sim \mathcal{D}} \left[ \nabla_w \pi(x; w)  (\pi(x; w) - \pi^\star(x)) \right].
\end{align}
Comparing the gradient of both losses, we first note that both require a forward and backward pass of $\pi$ to compute $\pi(x; w)$ and the gradient of the policy $\nabla_w \pi(x; w)$.
The exact Q-loss~\eqref{eq: gradient L_Q} additionally requires the gradient $\left.\nabla_u Q(x, u) \right|_{u = \pi(x_i; w)}$, which depends on $x$ and $u$.
Thus, we need to solve the OCP \eqref{ocp: q-function ocp} not only for each new sample $x$ but also for each new control $u$.
Note that we get the gradient via the multiplier when solving this OCP (cf. Lemma \ref{lemma: lagrangian multiplier}).

Since we might evaluate the loss \eqref{eq: exact q-loss} for any couple $(x, \pi(x;w))$ we need to guarantee feasibility of problem \eqref{ocp: q-function ocp} by introducing slack variables to soften the constraint, as presented in problem \eqref{ocp: discrete-time OCP}.

\begin{figure}
    \centering
    \includegraphics{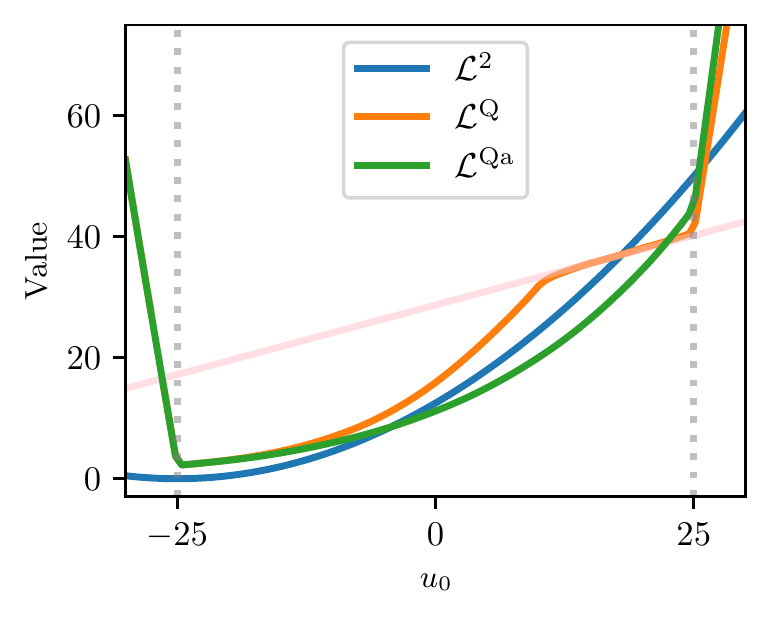}
    \caption{Comparison of the losses for the example presented in Section \ref{sec: example}, for a given initial state $\bar{x}_0 = (0.8, 0, \pi/4, 0)$ and $u_0^\star(\bar{x}_0) = -25$. The pink line corresponds to the gradient of $\mathcal{L}^Q$ at $\bar{u}_0 = 15$.}\label{fig: loss comparison}
\end{figure}

Finally, we remark that according to the properties of \eqref{ocp: q-function ocp}, the function $Q$ might be a piece-wise nonlinear and nonconvex function.
As a result of the nonconvexity in $u_0$, the tangent plane derived from the gradient of $Q$ at a given point $(\bar{x}_0, \bar{u}_0)$ might not be a lower bound for $Q(\bar{x}_0, u^\star_0)$, an example of this is given in Figure \ref{fig: loss comparison}.
This might slow down the minimization of the loss \eqref{eq: exact q-loss}.

In the next section, we address the computational complexity of the exact Q-loss issue by proposing a convex approximation of it.

\subsection{The Gauss-Newton Q-loss}
We propose a simplified loss which aims to alleviate both the computational burden related to $\mathcal{L}^\mathrm{Q}$ and the possible misleading gradients generated by the nonconvexity of $Q$.
The loss exploits the optimal control structure of $\mathcal{L}^\mathrm{Q}$ but it builds a quadratic programming approximation of the Q-function OCP \eqref{ocp: q-function ocp} around the optimal solution.

\assumption{
    Let us consider functions $L$ and $E$ in \eqref{cf: discrete-time OCP} being the square of vector-valued functions of the form $\Vert \bar{L} \Vert^2$ with $\bar{L}: \mathbb{R}^{n_x} \times \mathbb{R}^{n_u} \rightarrow \mathbb{R}^{n_y}$, $\Vert \bar{E} \Vert^2$ with $\bar{E}: \mathbb{R}^{n_x} \rightarrow \mathbb{R}^{n_e}$ respectively.
    Cost functions of this type appear frequently in optimal control since they represent tracking cost, most importantly they allow for a Gauss-Newton Hessian approximation~\cite{Verschueren2016}.
}

Given a sample $\bar{x}_0$ and a first control $\bar{u}_0$, we solve~\eqref{ocp: discrete-time OCP} and denote its solution as $\zeta =(\tilde{x}_0, \tilde{u}_0, \dots, \tilde{u}_{N-1}, \tilde{x}_N)$ and we use it as a linearization point for the quadratic approximation of problem~\eqref{ocp: q-function ocp} as follows
\begin{minipage}{\columnwidth}
\begin{flalign}
    Q_\mathrm{a}(\bar{x}_0, \bar{u}_0) \coloneqq && \nonumber
\end{flalign}
\begin{mini}[2]
    {\substack{x_0, u_0, \dots, \\ u_{N-1}, x_N}}{\sum_{k=0}^{N-1} \Vert L_{\mathrm{L}}(x_k, u_k; \tilde{x}_k, \tilde{u}_k) \Vert^2 + \Vert E_{\mathrm{L}}(x_N; \tilde{x}_{N}) \Vert^2}{\label{ocp: approx q-function OCP}}{}
    \addConstraint{x_0 - \bar{x}_0}{= 0}{}
    \addConstraint{u_0 - \bar{u}_0}{= 0}{}
    \addConstraint{x_{k+1} - f_\mathrm{L}(x_k, u_k; \tilde{x}_k, \tilde{u}_k)}{= 0}
    \addConstraint{h_\mathrm{L}(x_k, u_k; \tilde{x}_k, \tilde{u}_k)}{\leq 0,\; k=1,\dots,N-1}{}
    \addConstraint{r_\mathrm{L}(x_N; \tilde{x}_{N})}{\leq 0,}{}
\end{mini}
\end{minipage}

with $L_{\mathrm{L}}(x_k, u_k; \tilde{x}_k, \tilde{u}_k)$ being defined as the first order Taylor series of $L$ at $(\tilde{x}_k, \tilde{u}_k)$ as follows
\begin{align}
    L_{\mathrm{L}}(x_k, u_k; \tilde{x}_k, \tilde{u}_k) &= L(\tilde{x}_k, \tilde{u}_k)\\&+ \nabla_{x, u} L(\tilde{x}_k, \tilde{u}_k)^\top \left(\begin{bmatrix} x \\ u \end{bmatrix} - \begin{bmatrix} \tilde{x}_k \\ \tilde{u}_k \end{bmatrix}\right). \nonumber
\end{align}
The functions $f_{\mathrm{L}}, h_{\mathrm{L}}$ are defined in the same way, while for $E_{\mathrm{L}}, r_{\mathrm{L}}$ the linearization is done at $\tilde{x}_N$ and only with respect to $x$.
The function $Q_\mathrm{a}$ is now described by a convex piece-wise quadratic function.
\begin{lemma}
    If we use the optimal solution of \eqref{ocp: discrete-time OCP} as linearization point, i.e., set $(\bar{\bar{x}}_0, \bar{\bar{u}}_0)\coloneqq (\bar{x}_0, \pi^\star(\bar{x}_0))$, then the Gauss-Newton Q-loss is a convex distance function with $Q_\mathrm{a}(x, u) \geq Q_\mathrm{a}(x, \pi^\star(x))$, for every $x,u$.
\end{lemma}

We can introduce the approximate Q-loss as
\begin{align}
    \mathcal{L}^{\mathrm{Q_a}}(w) \coloneqq \E_{x \sim \mathcal{D}} \left[ Q_\mathrm{a}(x, \pi(x; w)) \right],
\end{align}
and its gradient $\nabla \mathcal{L}^{\mathrm{Q_a}}(w)$ is given by
\begin{align}\label{eq: gradient L_QP}
    \nabla \mathcal{L}^{\mathrm{Q_a}}(w) = \E_{x \sim \mathcal{D}} \left[ \nabla_w \pi(x; w) \left.\nabla_u Q_\mathrm{a}(x, u; \zeta)\right|_{u = \pi(x; w)} \right].
\end{align}
Compared to \eqref{eq: gradient L_Q} the function $\nabla \mathcal{L}^{\mathrm{Q_a}}$ requires the gradient of the QP problem, i.e., $\nabla_u Q_\mathrm{a}$ which is less expensive to compute.

\section{Numerical example}\label{sec: example}
\begin{figure}
    \centering
   \includegraphics{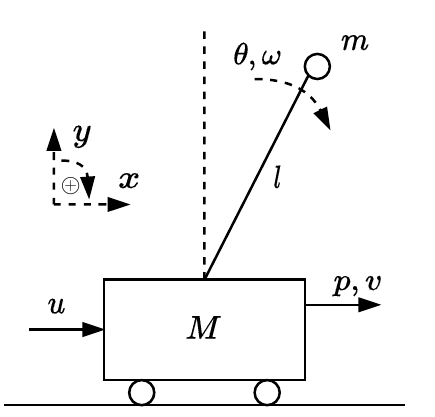}
    \caption{Schematic of the cart pole}\label{fig: cartpole}
\end{figure}

We show the effectiveness of the proposed losses against the standard $\ell_2$ loss on the example of the cart-pole.
The system is depicted in Figure \ref{fig: cartpole}.
The task is to control the system such that the rod stays in upright position and the cart stays at the center of the track.

By neglecting friction forces, the dynamics of the system are defined by the following equations
\begin{align}\label{eq: cartpole ode}
\begin{bmatrix}
    \dot{p}(t) \\ \dot{v}(t) \\ \dot{\theta}(t) \\ \dot{\omega}(t)
\end{bmatrix} =
\begin{bmatrix}
    v \\
    \frac{-m l \sin(\theta) \dot{\theta}^2 + m g \cos{\theta} \sin(\theta) + u}{M + m - m \cos^2(\theta)} \\
    \omega \\
    \frac{-m l \cos(\theta) \sin(\theta) \dot{\theta}^2 + (M + m) g \sin(\theta) + u \cos(\theta)}{l \cdot (M + m - m \cos^2(\theta))}
\end{bmatrix},
\end{align}
with $l = 0.8 \; (\mathrm{m}), m = 0.1 \; (\mathrm{kg})$, $M = 1 \; (\mathrm{kg})$, $g = 9.81 \; (\mathrm{m/s^2})$.
The system state is $x(t) = (p(t), v(t), \theta(t), \omega(t)) \in \mathbb{R}^4$, the control is $u(t) \in \mathbb{R}$.
We assume full state observability.

We use multiple shooting with $N = 20$ shooting intervals of $\Delta t = 0.05 \, \mathrm (s)$ and an RK4 integrator to discretize~\eqref{eq: cartpole ode} and obtain the discrete time OCP
\begin{mini}[2]
    {\substack{x_0, u_0, \dots, \\ u_{N-1}, x_N}}{\frac{1}{N} \left( \sum_{k=0}^{N-1} \begin{bmatrix} x_k \\ u_k \end{bmatrix}^\top \begin{bmatrix} S & 0 \\ 0 & R \end{bmatrix} \begin{bmatrix} x_k \\ u_k \end{bmatrix} + x_N^\top P x_N\right)}{\label{ocp: cartpole}}{}
    \addConstraint{x_{k+1}}{= f(x_k, u_k),}{\quad k=0, \dots, N-1}
    \addConstraint{\underline{x}}{\leq x_k \leq \bar{x}}{\quad k=0,\dots, N}
    \addConstraint{\underline{u}}{\leq u_k \leq \bar{u}}{\quad k=0,\dots, N-1}
    \addConstraint{x_0}{= \bar{x}_0.}{}
\end{mini}
with horizon length, $\bar{x} = (2, 4, \frac{\pi}{3}, 2), \underline{x} = -\bar{x}$ and $\bar{u} = 25, \underline{u} = -\bar{u}$.
The weight matrices in the cost function are $S = \mathrm{diag}(0.25, 0.025, 0.25, 0.025)$, $R = 0.0025$ and $P$ corresponds to the solution of the discrete algebraic Riccati equation for the system linearized at $\bar{x}=(0,0,0,0), \bar{u}=0$.
In order to guarantee feasibility during training the box constraints on $(x, u)$ are softened via slack variables which are penalized in the cost with the weights $Z = \Delta t \cdot (50, 5, 50, 5, 500), z =\Delta t \cdot (0.5, 0.05, 0.5, 0.05, 5000) $ for the path constraints and $Z_e = (50, 5, 50, 5), z_e = (0.5, 0.05, 0.5, 0.05) $ for the terminal ones.

By modifying the OCP formulation \eqref{ocp: cartpole} according to \eqref{ocp: q-function ocp} and \eqref{ocp: approx q-function OCP} we obtain the Q-function OCP and the approximate Q-function OCP, respectively.

The formulation and solution of the OCP is carried out in \texttt{acados}~\cite{Verschueren2021} via its Python interface.
In order to speed up interactions with the OCP solver, we have used the compiled Cython interface for the solver objects.
Every computation runs exclusively on one CPU thread, on a Linux Ubuntu 20.04 server with Intel Xeon E5-2687W @3.1 GHz, 16 cores and 32 GB RAM.

\subsection{Training Setup}
For approximating the MPC policy we use feed-forward Neural Networks with ReLU activation functions for the hidden layers.
Additionally, after the last linear layer, we apply a $\tanh$ activation function, which bounds the output of the policy such that we fulfill the box constraints for the controls of the cart-pole OCP formulation \eqref{ocp: cartpole}.
We optimize the Neural Networks by doing mini-batch stochastic gradient descent with the Adam optimizer \cite{Kingma2014}.

The next important decision is on what states we want to imitate, or more specifically, on which state distribution $\mathcal{D}$ we want to minimize our imitation loss.
For this, we first sample an initial state uniformly from $\mathcal{U}(\alpha \cdot \underline{x}, \alpha \cdot \bar{x})$ with $\alpha = 0.3$.
We sample and discard until the drawn initial states generate optimal open-loop trajectories without constraint relaxation.
Starting from the initial state, we then do a rollout with the Dagger algorithm as described in \cite{Ross2011}.
Instead of only following the expert MPC policy during the rollout, we use a mixture policy that randomly applies either the control of the expert MPC policy or our currently learned policy $\pi(x ; w)$.
This is done in order to generate samples that better match the distribution that we will encounter when applying the final learned policy.
In more detail, we train the Neural Network for $2000$ updates and every $20$ updates collect additional samples by doing a rollout for $50$ steps.
The mixture coefficient between the expert policy and the learned policy is linearly scaled down over the training time from $1$, we only use the expert, to $0$, we only use the learned policy.
\subsection{Evaluation}
For evaluation, we train each algorithm for $10$ different random seeds.
For the initial states we either sampled from a uniform distribution with easier initial states $\alpha = 0.2$, the same distribution as during training $\alpha = 0.3$ or harder initial states with $\alpha = 0.4$.
We do this to see how the performance on easier and harder initial states differ and also test the generalization capabilities of the different losses.
We generate one fixed test dataset of $2000$ initial states for all algorithms and seeds, with the same sampling procedure as during training.
For each initial state, we then do a rollout of $50$ steps for each learned policy.

We evaluate the performance of the final policies with two metrics:
(1) The average rollout cost: We sum up all stage costs and slack variable costs of one rollout and then average over all rollouts and random seeds.
(2) The average violation ratio, which is the ratio of rollouts that violated a constraint over all rollouts, averaged over all random seeds.
Additionally, we only consider the $0.99$ or $0.9$ quantile to robustify our estimates against outliers, when the policy fails to stabilize the system.

\subsection{Hyperparameters}
To allow for a better comparison, we do a hyperparameter search in form of a grid search over the network depth $\{1, 2, 3\}$, the network width $\{64, 128, 256\}$ and the learning rate $\{10^{-5}, 10^{-4}, 10^{-3}, 10^{-2}\}$.
We used a fixed batch size of $32$.
We use the same hyperparameters for the Gauss-Newton Q-loss as for the exact Q-loss.
The search objective is the average rollout cost over the same initial state dataset as described in the previous section, but only using $3$ random seeds.

\subsection{Results}

\begin{table}
	\vspace{0.3cm}
    \centering
    \ra{1.2}
    \caption{Loss performance comparison}
    \begin{tabular}{@{}lcccc@{}} \toprule
        Loss & $\alpha$ & Quantile & Avg. Cost & Violation ratio \\ \midrule
        \multirow{3}*{ $\mathcal{L}^2$ }                & 0.2   &  0.99 & $0.866 \pm 0.013$   & $0.054 \pm 0.042$  \\
                                                        & 0.3   &  0.99 & $2.267 \pm 0.250$   & $0.225 \pm 0.064$  \\
                                                        & 0.4   &  0.90  & $3.229 \pm 0.385$   & $0.299 \pm 0.064$  \\ \midrule
        \multirow{3}*{ $\mathcal{L}^{\mathrm{Q}}$}      & 0.2   &  0.99  & $0.919 \pm 0.093$   & $0.005 \pm 0.012$  \\
                                                        & 0.3   &  0.99  & $2.308 \pm 0.209$   & $0.090 \pm 0.049$  \\
                                                        & 0.4   &  0.90  & $3.267 \pm 0.365$   & $0.147 \pm 0.059$  \\ \midrule
        \multirow{3}*{ $\mathcal{L}^{\mathrm{Q_a}}$}    & 0.2   &  0.99  & $0.884 \pm 0.019$   & $0.002 \pm 0.001$  \\
                                                        & 0.3   &  0.99  & $2.179 \pm 0.081$   & $0.088 \pm 0.024$  \\
                                                        & 0.4   &  0.90  & $3.076 \pm 0.150$   & $0.154 \pm 0.025$  \\ \midrule
        \multirow{3}*{ $\pi^\star$ MPC}                 & 0.2   &  0.99  & $0.795 \pm 0.833$   & $-$       \\
                                                        & 0.3   &  0.99  & $1.888 \pm 2.027$   & $-$        \\
                                                        & 0.4   &  0.90  & $2.604 \pm 2.373$   & $-$       \\
        \bottomrule
    \end{tabular}\label{tab: performance comparison}
	\vspace{-0.3cm}

\end{table}

In the following, we present the experimental results.
In Table \ref{tab: performance comparison}, we compare the performance of the different algorithms.
For the average rollout cost, we see that the exact Q-loss performs slightly worse than BC, whereas the Gauss-Newton Q-loss performs significantly better for the harder initial state dataset $\alpha = 0.4$.
However, for the harder examples the gap to the original MPC policy is also significantly larger.
One explanation is that for harder examples the non smoothness of the optimal control policy increases, making it harder for the network to approximate the MPC, especially for BC.

For the average violation ratio, we see that the exact Q-loss and the Gauss-Newton Q-loss perform significantly better, than BC with $\mathcal{L}^2$.
This can be attributed to the fact that the exact Q-loss and the Gauss-Newton Q-loss contain constraint satisfaction and do not rely on a surrogate loss.

\begin{figure}
    \centering
    \includegraphics[width=\columnwidth]{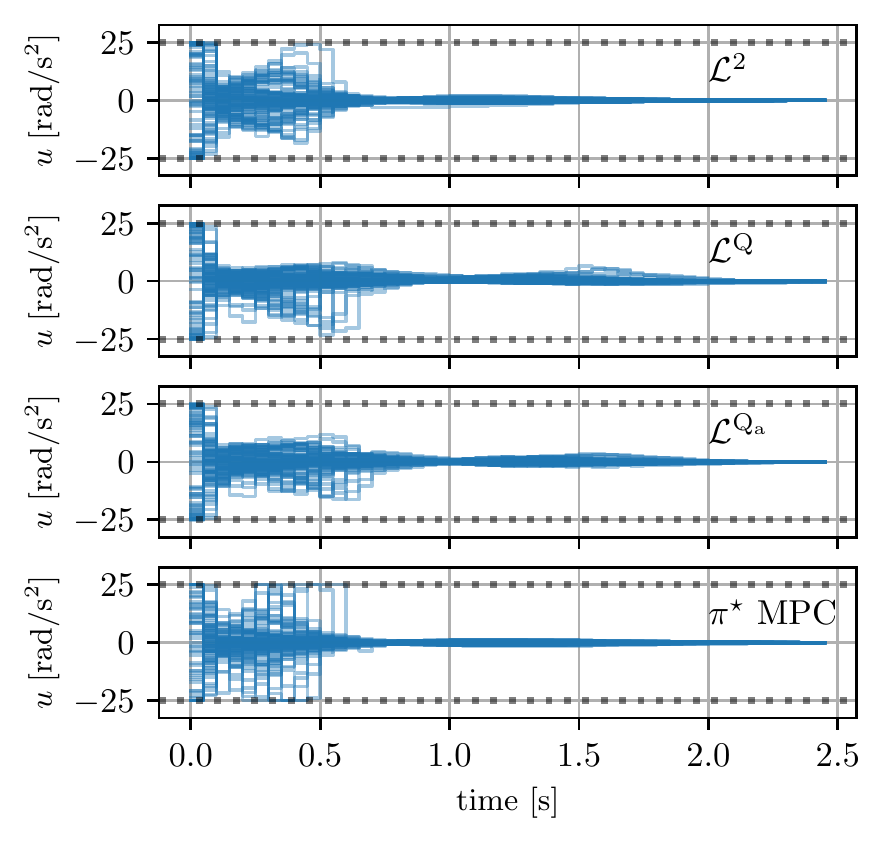}
    \caption{Representative 100 rollouts for $\alpha=0.3$ for one network.}\label{fig: rollouts}
\end{figure}
In Figure \ref{fig: rollouts}, we compare the controls of the policies learned with the different algorithms on exemplary rollouts of a trained Neural Network for one seed.
We see that the original MPC shows a very non smooth control signal, which the BC loss $\mathcal{L}^2$ tries to imitate.
The policies corresponding to the exact Q-loss and the Gauss-Newton Q-loss show a smoother control signal, that deviates more from the original MPC.
This can be explained by the fact that the proposed losses directly optimize (approximate) versions of the OCP, thus finding their own trade-off between feasibility and optimality.

\begin{table}
    \centering
    \ra{1.2}
    \caption{Average gradient computation speed}
    \begin{tabular}{@{}lccc@{}} \toprule
        Loss                    & $\mathcal{L}^2$ & $\mathcal{L}^{\mathrm{Q}}$ &  $\mathcal{L}^{\mathrm{Q_a}}$ \\ \midrule
        Speed (batch/second)    & $151.78$          & $4.12$                    & $15.23$\\ \bottomrule
    \end{tabular}\label{tab: gradient computation speed}
	\vspace{-0.3cm}

\end{table}

In Table \ref{tab: gradient computation speed}, we compare the average speed for computing the gradient on a batch (with batch size $32$) for $300$ iterations.

\section{Conclusions}
In this paper, we have presented a new loss for Imitation Learning from MPC based on the underlying OCP.
This loss allows the learned policy to directly minimize the OCP performance objectives and constraint satisfaction.
Compared to standard losses, the Q-loss evaluation requires the solution of a possibly nonlinear and nonconvex optimization problem for each new sample, resulting in demanding computational effort.
We suggest to mitigate this issue by a second Q-loss based on the Gauss-Newton approximation of the associated OCP, therefore its evaluation requires the solution of a convex quadratic program.
Finally, we have compared the policy learned using the Q-losses against Behavioral Cloning, on the control of a constrained nonlinear system.
On this example, the Q-loss-based policies achieve significantly lower constraint violations and comparable closed-loop costs.
In the future, we aim to test the losses on more complex examples and combine them with Reinforcement Learning.

\bibliography{bib4arxiv}

\end{document}